\documentclass[10pt,twocolumn,letterpaper]{article}

\usepackage{iccv}
\usepackage{times}
\usepackage{epsfig}
\usepackage{graphicx}
\usepackage{amsmath}
\usepackage{amssymb}

\usepackage{array}
\usepackage{booktabs}
\usepackage{xcolor}
\usepackage[numbers]{natbib}
\usepackage{multirow}
\usepackage{soul}
\usepackage{subcaption}
\usepackage{algpseudocode}
\usepackage{algorithm}
\usepackage{algorithmicx}
\usepackage{booktabs}
\usepackage{widetable}
\usepackage{adjustbox}
\usepackage[inline]{enumitem}
\usepackage{bm}
\usepackage[mathscr]{euscript}
\usepackage{makecell}
\usepackage[hyphens]{url}
\usepackage{pifont}

\usepackage{diagbox}
\usepackage{color}
\usepackage{colortbl}
\usepackage{mathtools}

\usepackage[skip=2pt,font=small]{caption}

\newcommand{\ourModelName}{pair-wise self-consistency learning}
\newcommand{\ourModelNameCap}{Pair-Wise Self-Consistency Learning}
\newcommand{\ourModelAbbr}{PCL}
\newcommand{\ourGeneratorName}{inconsistency image generator}
\newcommand{\ourGeneratorNameCap}{Inconsistency Image Generator}
\newcommand{\ourGeneratorAbbr}{I2G}

\newcommand{\xhdr}[1]{\vspace{5pt} \noindent {\textbf{#1}}}
\newcommand{\quotes}[1]{``#1''}

\usepackage[breaklinks=true,bookmarks=false,colorlinks=true]{hyperref}

\iccvfinalcopy 

\def\iccvPaperID{1521} 

\ificcvfinal\fi

\begin{document}

\title{Learning Self-Consistency for Deepfake Detection}

\author{
    Tianchen Zhao\thanks{Currently at University of Michigan, Ann Arbor. The work was conducted while at Amazon/AWS AI.}
    \hspace{0.3cm} Xiang Xu
    \hspace{0.3cm} Mingze Xu
    \hspace{0.3cm} Hui Ding
    \hspace{0.3cm} Yuanjun Xiong
    \hspace{0.3cm} Wei Xia \\ [.5ex]
    Amazon/AWS AI \\ [.5ex]
    {\tt\small ericolon@umich.edu, \{xiangx,xumingze,huidin,yuanjx,wxia\}@amazon.com}
}

\maketitle
\ificcvfinal\thispagestyle{empty}\fi

\begin{abstract}
We propose a new method to detect deepfake images using the cue of the source feature inconsistency within the forged images. It is based on the hypothesis that images' distinct source features can be preserved and extracted after going through state-of-the-art deepfake generation processes. We introduce a novel representation learning approach, called \ourModelName~(\ourModelAbbr), for training ConvNets to extract these source features and detect deepfake images. It is accompanied by a new image synthesis approach, called~\ourGeneratorName~(\ourGeneratorAbbr), to provide richly annotated training data for~\ourModelAbbr. Experimental results on seven popular datasets show that our models improve averaged AUC over the state of the art from 96.45\% to 98.05\% in the in-dataset evaluation and from 86.03\% to 92.18\% in the cross-dataset evaluation.
\end{abstract}

\section{Introduction}
\label{sec:intro}


Deepfakes are synthetic media in which the identity or expression of a target subject is replaced by that of another source subject. 
They are predominantly generated by image stitching, which includes face detection, warping, and blending.
Attacks using deepfakes have caused a significant amount of negative social impact, and also motivated methods to detect these forged videos.
Most of these defense methods~\cite{zhou-cvprw17,li-wifs18,nguyen-btas19,wang-arxiv19,tolosana-arxiv20,dang-cvpr20,li-cvpr20oral,kumar-arxiv20,lima-arxiv20} target detecting suspicious artifacts left in the stitching process,
such as eye blinking~\cite{li-wifs18}, face warping~\cite{li-cvprw19}, blending boundaries~\cite{li-cvpr20oral}, and fake prototypes~\cite{trinh2021wacv}.
In the wake of these defenders, forgery techniques are also evolving on reducing these artifacts to avoid detection, forming an enduring arms race.

\begin{figure}[t!]
\centering
\includegraphics[width=1.0\linewidth]{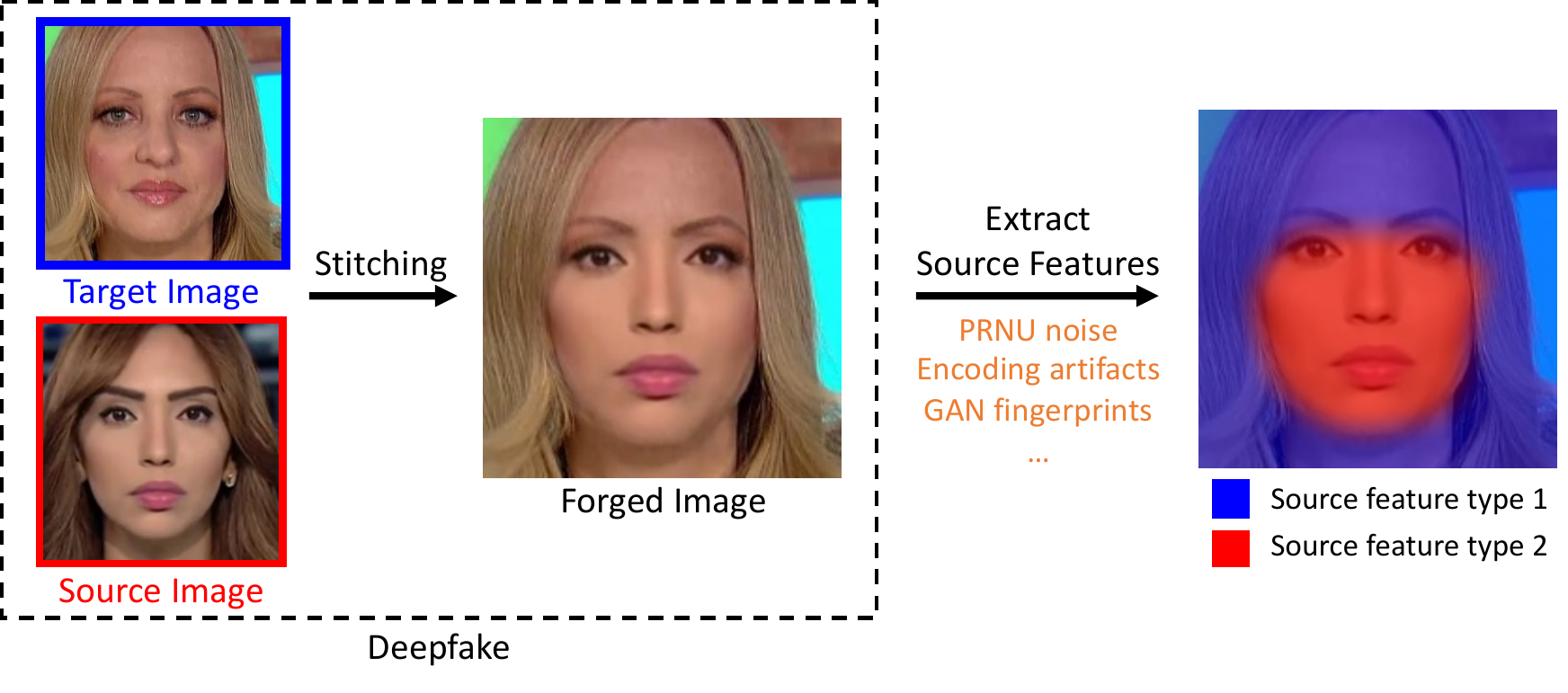}
\caption{
The forged image is generated by stitching target and source images.
We hypothesize that each of them carries distinct source features that can uniquely identify their sources.
Therefore, the forged image contains different source features at different locations, whereas those of a pristine image must be consistent across all positions.
By extracting the local source features and measuring their self-consistency, we can detect forged images.
}
\vspace{2pt}
\label{fig:overview}
\end{figure}


In this paper, we propose a new method to detect deepfakes generated by stitching-based methods. 
Unlike other methods focusing on detecting artifacts described above, our approach uses the cue of \emph{inconsistency of source features within the forged images}.
Conceptually, images carry content-independent~\cite{huh-eccv18}, spatially-local information that can uniquely identify their sources. We call them the \textit{source features}. 
They could come from either imaging pipelines (\eg, PRNU noise~\cite{lukas-ivcp05}, specifications~\cite{huh-eccv18}), encoding approaches (\eg, JPEG compression patterns~\cite{barni-vci17}, compression rates) or image synthesis models~\cite{yu-iccv19}.
We hypothesize that these source features are still preserved after the modified image having gone through the state-of-the-art deepfake generation processes~\cite{li-cvpr20_2,dolhansky-arxiv20,jiang-cvpr20,li-cvpr20,perez-acm03,naruniec-egsr20}.
Therefore, a forged image would contain different source features at different positions, whereas those of a pristine image must be consistent across all positions.
By extracting the local source features and measuring their self-consistency, we can detect forged images.


Specifically, we use a convolutional neural network (ConvNet) to extract source features in the form of down-sampled feature maps. 
Each feature vector represents the source features of a corresponding location in the input image.
To train this ConvNet,
we introduce a novel representation learning method, called \ourModelName~(\ourModelAbbr), which uses the consistency loss for supervision.  
We calculate the cosine similarity between very pairs of feature vectors in the source feature map, and compute the consistency loss on
all pairs according to whether their corresponding image locations come from the same source image.
That is, we penalize the pairs that refer to locations from the same source image for having a low-similarity score and those from different source images for having a high-similarity score.
We attach a non-linear binary classifier on the learned source feature map to perform the deepfake detection. We train it with an additional loss to produce the image-level real vs. fake labels.


The consistency loss in~\ourModelAbbr~needs pixel-level annotation about whether a location has been modified. 
It is generally not available in deepfake detection datasets, on which the re-annotation could be laborious and error-prone. 
We use synthesized data generated from \ourGeneratorName~(\ourGeneratorAbbr) to tackle this issue. 
It generates forged images following the latest techniques in deepfake generation methods.
To save computational cost and enable online generation, \ourGeneratorAbbr~only stitches together pristine source and target images instead of the synthesized ones from deep networks.
We randomly sample the forgery mask for stitching during generation, which becomes the pixel-level annotation we need for~\ourModelAbbr. 
Experimental results show that, although using a simplified generation process, the models learned with synthesized data from~\ourGeneratorAbbr~effectively extract discriminative source features in both pristine and deepfake images.


We evaluate~\ourModelAbbr~on seven recent deepfake detection datasets and observe superior detection accuracy.
Following the in-dataset evaluation, our method achieves the AUC scores of 99.79\%, 99.98\%, and 94.38\% on FF++, CD2, and DFDC-P datasets, respectively.
Because~\ourModelAbbr~uses the cue of source feature inconsistency which is less taken care of by current deepfake generation methods, we conjecture that a model trained with~\ourModelAbbr~on one dataset could effectively detect deepfakes generated by methods not seen in this dataset.  
To verify this, we adopt the cross-dataset evaluation protocol introduced in~\cite{li-cvpr20oral} to test our models and observe affirmative results.
We achieve the AUC scores of 99.11\%, 99.07\%, 99.41\%, 98.30\%, and 90.03\% on FF++, DFD, DFR, CD1, and CD2 datasets, respectively.
We further visualize the consistency map of the learned source features on both real and fake images. We observe the consistency maps can lead to localization of the modified region.

It is worth noting that, as the race between forgers and defenders continues, the cue of source feature inconsistency can be negated. 
It can be done by either using entire face synthesis techniques~\cite{karras-arxiv19,liu-cvpr19,choi-cvpr20} that directly output the whole fake image, such as GAN, or future development of stitching methods that completely removes or ambiguates the source feature. 
However, the state-of-the-art deepfake generation methods have not yet adopted these techniques. 
Thus the effectiveness of our method on detecting deepfake images should only be evaluated on the images generated by existing deepfake detection methods, as depicted in deepfake detection datasets we used~\cite{roessler-iccv19,dufour-dfd,li-cvpr20_2,dolhansky-arxiv20,dolhansky-arxiv19,jiang-cvpr20}.

\section{Related Work}
\label{sec:related_work}

\begin{figure*}[t]
\centering
\includegraphics[width=1.0\linewidth]{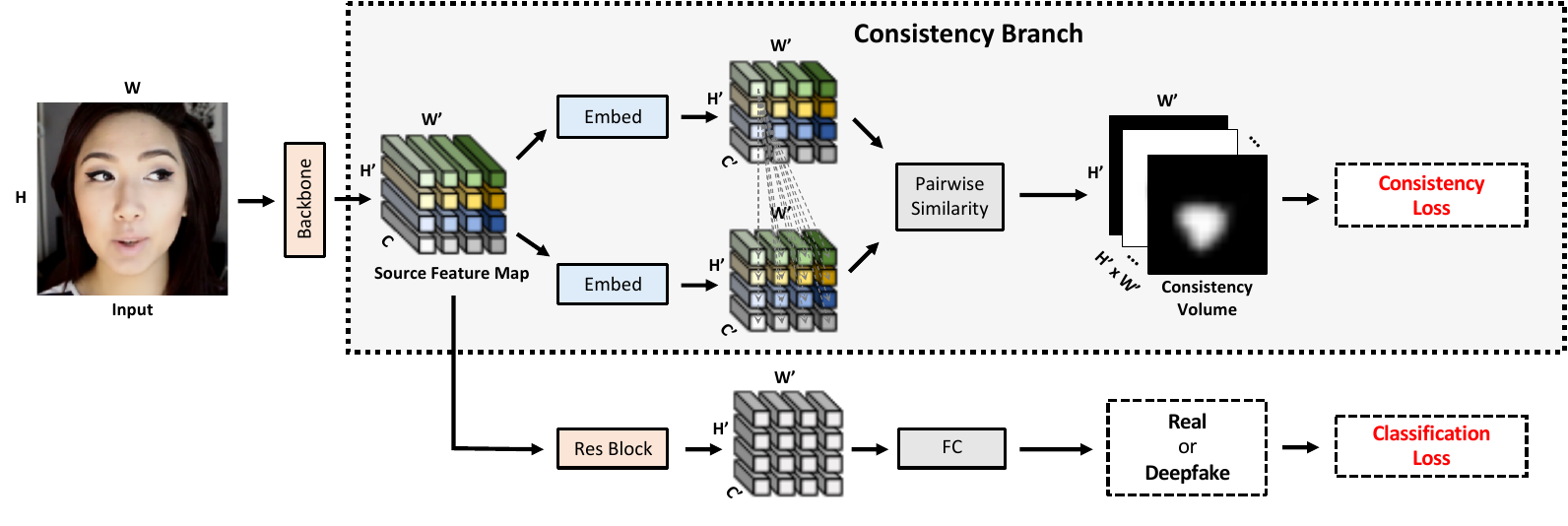}
\caption{
\textit{Visualization of \ourModelAbbr~architecture.}
The consistency branch focuses on measuring the consistencies of image patches according to their source features.
A classification branch is applied after the source feature map and predicts the binary score for deepfake detection.
}
\vspace{-7pt}
\label{fig:network}
\end{figure*}

\vspace{-5pt} \xhdr{Deepfake Generation.}
There are four common types of deepfake~\cite{tolosana2020deepfakes}: entire image synthesis, modification of facial attribute or expression, and face identity swap.
3D models \cite{thies-cvpr16}, AutoEncoders \cite{tewari2017mofa,vahdat2020nvae}, or Generative Adversarial Networks \cite{karras-cvpr19,karras-arxiv19,liu-cvpr19,choi-cvpr20} are used to generate the fake image segment, which is then blended back to the original image.

\xhdr{Deepfake Detection.}
To detect the whole image synthesis, recent research~\cite{yu-iccv19,neves-jstsp20,huang-arxiv20,wang-cvpr20} observes that GAN-generated images contain specific cues that can be easily detected, and the trained models have exhibited good generalization ability across different synthesis methods.
To support the research on detecting the other types of face manipulations, several deepfake datasets are released~\cite{yang-icassp18,korshunov-arxiv18,roessler-iccv19,dufour-dfd,li-cvpr20_2,dolhansky-arxiv20,jiang-cvpr20,zi-acm20} and countermeasures have been introduced. 
FakeSpotter~\cite{wang-arxiv19} is proposed with a layer-wise neuron behavior for fake face detection. 
Recurrent neural networks~\cite{sabir-cvprw19} and various types of 3D ConvNets~\cite{lima-arxiv20} are utilized to detect the manipulation artifacts across the video frames.
However, binary classifiers are criticized for their interpretability, and several localization methods are introduced through either multi-task learning~\cite{nguyen-btas19} or attention-based mechanisms~\cite{dang-cvpr20,zhao-cvpr21}.
To improve the generalization ability,
DSP-FWA~\cite{li-cvprw19} and Face X-ray~\cite{li-cvpr20oral} also make their data generation pipeline and the latter focuses on predicting the blending boundaries in fake video frames.

Our approach also lies in this line but has several key differences.
First, from the methodology perspective,
we focus on detecting deepfakes by using a less attended cue of inconsistency of source features within the forged images.
Second, from the network design perspective, our consistency predictor only contains a few parameters and can serve as a plugin module upon any common backbones.

\xhdr{Consistency Learning.} The concept of inconsistency has been studied in the image forensic literature~\cite{mayer-sp20},
where similarity scores are computed among image patches~\cite{mayer-sp20,mayer-ifs19,huh-eccv18,mayer-icassp18,zhou-cvprw17,bondi-spl16}.
Zhou~\etal~\cite{zhou-cvprw17} propose a two-stream network to detect
both tampered faces and low-level inconsistencies,
but the training
requires
steganalysis feature extraction.
Huh~\etal~\cite{huh-eccv18} use a Siamese network to predict the metadata inconsistency by iteratively
comparing random patches from different raw images.
Nirkin~\etal~\cite{nirkin-pami20} use signals from the proposed face identification and context recognition networks to detect deepfakes.

In this paper, we introduce consistency learning to deepfake detection and propose an end-to-end learning architecture that estimates the image self-consistency with one forward pass, while capturing the internal relations among patches within an image.
Besides, instead of using only raw images, we design \ourGeneratorAbbr~to address several challenges to fit face forgery detection better by supplying \ourModelAbbr~with training images that are finely stitched from multiple sources.

\section{Our Approach}
\label{sec:our_approach}

Given an input image, our goal is to detect if the identity or expression of the subject is
replaced with that of another subject.
Observing that deepfakes are stitched by images from different sources with distinct source features, we explore learning effective and robust representations for deepfake detection by measuring the source feature consistency within the image.
More specifically, we propose a multi-task learning architecture, as shown in Fig.~\ref{fig:network}.
The consistency branch is optimized to predict a consistency map for each image patch, indicating its source feature consistencies with all others.
The classification branch is applied to the source features and outputs binary labels for evaluation purposes.
The model is trained on both consistency and classification loss, with annotations supplied by~\ourGeneratorAbbr.

\subsection{\ourModelNameCap~(\ourModelAbbr)}
\label{sec:our_approach:pcl}

\vspace{-5pt} \xhdr{The consistency branch}
computes the pairwise similarity scores of all possible pairs of local patches in an image and predicts a 4D consistency volume $\widehat{\textbf{V}}$.
Given a pre-processed video frame $X$ of size $H$\texttimes$W$\texttimes$3$ as input, we first feed it into the backbone and extract the source feature $F$ of size $H'$\texttimes$W'$\texttimes$C$ from an intermediate convolution layer, where $H'$, $W'$, and $C$ are height, width, and channel size, respectively.
For each patch $\mathcal{P}_{h,w}$ in the source feature map, we compare it against all the rest to measure their feature similarities, and obtain a 2D consistency map $\widehat{M}^{\mathcal{P}_{h,w}}$ of size $H'$\texttimes$W'$ of consistency scores in the range of $[0, 1]$, where the superscript indicates the position of the base patch.
To be specific, for any pair of patches $\mathcal{P}_{i}$ and $\mathcal{P}_{j}$, we compute their dot-product similarity~\cite{wang-cvpr18} using their extracted feature vector $f_i$ and $f_j$, both of size $C$, to estimate their consistency score:
\begin{align}
    s(f_i, f_j) = \sigma\left(\frac{\theta(f_i)\theta(f_j)}{\sqrt{C'}}\right),
\end{align}
where $\theta$ is the embedding function, realized by
$1$\texttimes$1$
convolutions, $C'$ is the embedding dimension, and $\sigma$ is the Sigmoid function.
We iterate this process over all patches $\{\mathcal{P}_{h,w} | 1 \leq h \leq H', 1 \leq w \leq W'\}$ in the source feature map, and finally get the 4D consistency volume $\widehat{\textbf{V}}$ of size $H'$\texttimes$W'$\texttimes$H'$\texttimes$W'$.
To easier interpret and provide visualization clues about the region of modification, we fuse the 4D consistency volume $\widehat{\textbf{V}}$ over all patches and generate a 2D global heatmap $\widehat{\mathcal{M}}$ of size $H'$\texttimes$W'$. We up-sample $\widehat{\mathcal{M}}$ to $\widehat{\mathscr{M}}$ of size $H$\texttimes$W$ to match the input size for visualization.

\xhdr{The optimization of consistency branch}
requires the 4D \quotes{ground truth} consistency volume $\textbf{V}$.
Given the mask $\mathscr{M}$ of size $H$\texttimes$W$ indicating the manipulated region of input $X$, we first create its coarse version $\mathcal{M}$ matching the size of $H'$\texttimes$W'$ through bi-linear down-sampling.
We obtain the ground truth 2D consistency map $M^{\mathcal{P}_{h,w}}$ for the $(h, w)$-th patch by computing the element-wise difference between its own value $\mathcal{M}_{h,w}$ and all others,
\begin{align}
    M^{\mathcal{P}_{h,w}} = 1 - \left|\mathcal{M}_{h,w} - \mathcal{M}\right|,
    \label{eq:heatmap_groundtruth}
\end{align}
where $\mathcal{M}_{h,w}$ is the scalar value in position $(h,w)$, and $M^{\mathcal{P}_{h,w}}$ is in size of $H'$\texttimes$W'$.
For each entry of $M^{\mathcal{P}_{h,w}}$, a value close to $1$ denotes the two patches are consistent, and close to $0$ otherwise.
To obtain the ground truth 4D global map $\textbf{V}$, we compute $M^{\mathcal{P}_{h,w}}$ for all patches.
Note that $\textbf{V}$ of a pristine image should be $\textbf{1}$, a 4D volume in which all values are equal or close to one.

We use the binary cross-entropy (BCE) loss to supervise the consistency prediction over the 4D consistency volume $\widehat{\textbf{V}}$, and more formally,
\begin{align}
    \mathcal{L}_{\textit{\ourModelAbbr}} =
    \frac{1}{N} \sum_{h,w,h',w'} \text{BCE}(\textbf{V}_{h,w,h',w'}, \widehat{\textbf{V}}_{h,w,h',w'}),
\label{eq:pair_loss}\end{align}
where $h$ and $h'$
$\in \{1, 2, ..., H'\}$,
$w$ and $w'$
$\in \{1, 2, ..., W'\}$,
and $N$ equals to $H'$\texttimes$W'$\texttimes$H'$\texttimes$W'$.

The consistency branch learns the representations that predict the self-consistency of the input according to their source features,
which by our claim could significantly benefit the deepfake detection in both effectiveness and robustness.
Nevertheless, these features cannot directly make inferences for evaluation purposes.

\xhdr{The classification branch}
is thus applied after the source feature map to predict if the input is real or fake.
More specifically, the extracted source feature is fed into another convolution operation. A global average pooling and fully-connected layer are built after that as the classifier, which outputs the probability score for the input of being real or fake.
We use the two-class cross-entropy (CE) loss $\mathcal{L}_{\textit{CLS}}$ to supervise the training in the classification branch.

The overall loss function of our model is as follows:
\begin{align}
    \mathcal{L} = \lambda \mathcal{L}_{\textit{\ourModelAbbr}} + \mathcal{L}_{\textit{CLS}},
\label{eq:total_loss}
\end{align}
with hyper-parameter $\lambda$.
The ablation study in Section~\ref{sec:exp:ablation_studies} suggests that a choice of large $\lambda$ value significantly improves the performance.
This observation demonstrates that the representations learned from the consistency branch play a dominant role in the success.

\subsection{\ourGeneratorNameCap~(\ourGeneratorAbbr)}
\label{sec:our_approach:i2g}

Training \ourModelAbbr~requires patch-level annotations of the manipulated regions, which is not always available in current existing datasets.
To provide this training data, we propose \textit{\ourGeneratorName~(\ourGeneratorAbbr)} to generate \quotes{self-inconsistent} images from the pristine ones, along with the ground truth masks $\mathscr{M}$ discussed in Section~\ref{sec:our_approach:pcl}.
To guarantee sufficient amount and diversity of the training data with least efforts, \ourGeneratorAbbr~reduces the computational cost by replacing facial image synthesis using the GAN or VAE ~\cite{li-cvpr20_2,dolhansky-arxiv20,jiang-cvpr20,li-cvpr20} with real images. It follows that \ourGeneratorAbbr~can support dynamic data generation on CPU during the training and be utilized as a part of the data augmentation for deepfake detection.

Similar self-supervised approaches~\cite{li-cvprw19,li-cvpr20oral,naruniec-egsr20,dolhansky-arxiv20} have been studied in other tasks or methods
for deepfake detection.
\ourGeneratorAbbr~particularly addresses several challenges to fit \ourModelAbbr~better.
\textbf{First}, because face images have some strong structural bias, stitching with the face hull regions may create undesired correlations between the source feature inconsistency and the face boundary.
\ourGeneratorAbbr~uses elastic deformation~\cite{ronneberger-icm15} to improve the variety of the mask $\mathscr{M}$ and thereby eliminates those spurious correlations.
\textbf{Second}, because attackers will intentionally try to remove source features to make deepfake images more realistic, \ourModelAbbr~needs to make use of source features that are not vulnerable to these approaches. 
\ourGeneratorAbbr~randomly selects one from an exhaustive set of blending methods in data generation so the representation learned by \ourModelAbbr~can be robust to source feature removal attempts.
\textbf{Third}, we expect the learned representation to generalize to a wide range of sources, even unseen ones during training. \ourGeneratorAbbr~adds image augmentation to the generation process to achieve this goal. The augmentation methods include JPEG compression, Gaussian noise/blur, brightness contrast, random erasing, and color jittering.

\begin{figure}[t]
\centering
\includegraphics[width=1.0\linewidth]{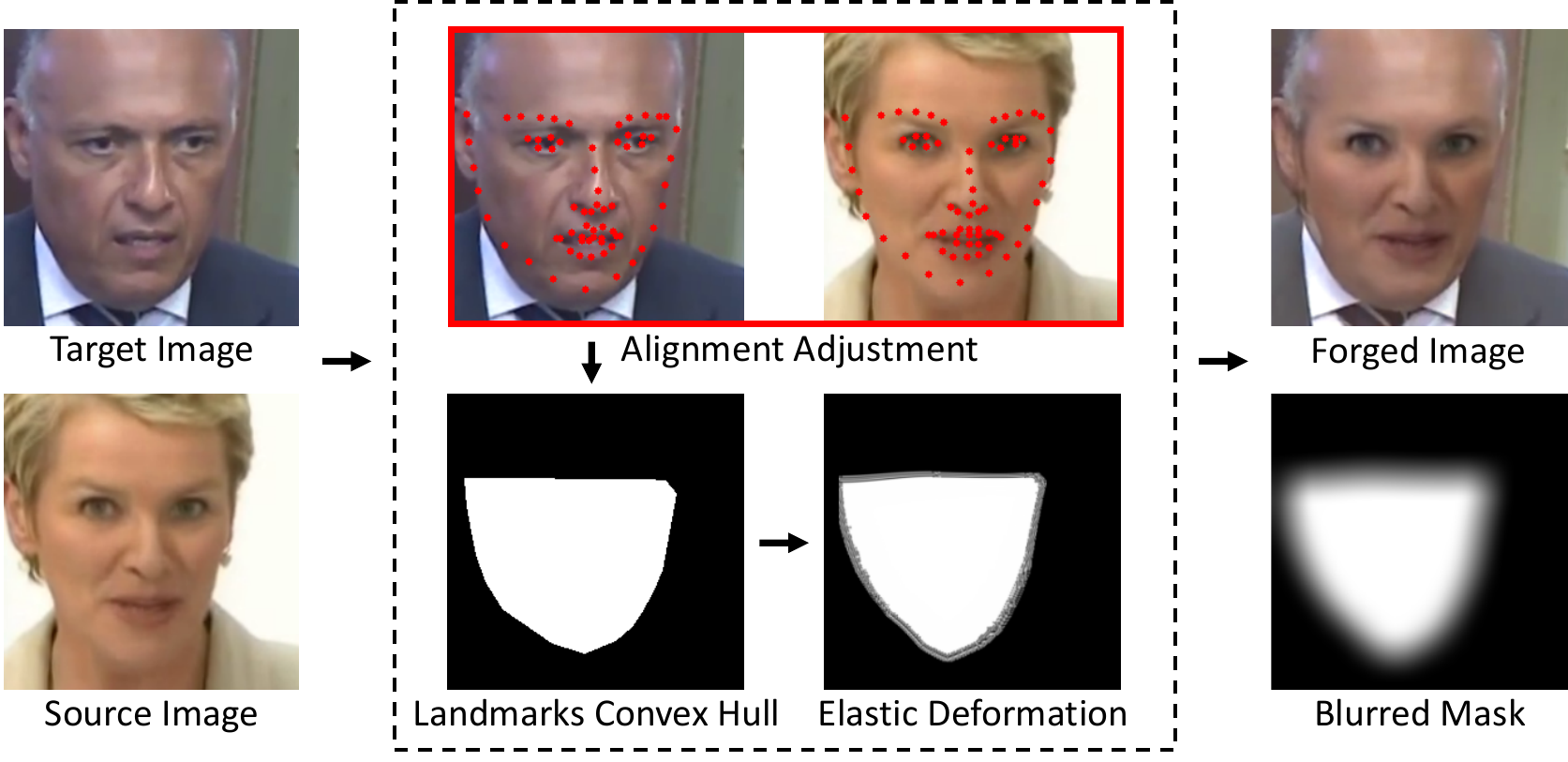}
\caption{
\textit{Illustration of the workflow of \ourGeneratorAbbr.}
For each source and target image pair, a morphed mask is generated by taking a convex hull of the landmarks followed by elastic deformation and Gaussian blur. The masked region of the target image is replaced by that of the source with blending techniques~\cite{perez-acm03,naruniec-egsr20}.
}
\label{fig:data_augmentation}
\end{figure}

\begin{algorithm}[t!]
    \caption{\ourGeneratorNameCap~(\ourGeneratorAbbr)}
    \hspace*{\algorithmicindent} \textbf{Input}: Target video frame $X^t$ of size $(H,W,3)$. \\
    \hspace*{\algorithmicindent} \textbf{Output}: Generated video frame $X^g$ and mask $\mathscr{M}$. \\
    \hspace*{\algorithmicindent} \textbf{Landmark Detector} $\mathcal{K}:
    \mathbb{R}^{H \times W \times 3}\rightarrow \mathbb{R}^{68 \times 2}$.
    \begin{algorithmic}[1]
        \State Get the video frame $X^t$ and its landmarks $\mathcal{K}(X^t)$.
        \State Find a random source frame $X^s$ of different ID, which
        satisfies $\lVert \mathcal{K}(X^t)-\mathcal{K}(X^s) \rVert_2<\epsilon$
        for threshold $\epsilon>0$.
        \State Align $X^s$ to $X^t$ using landmarks.
        \State Compute convex hull $\mathcal{H}$ of $\mathcal{K}(X^t)$.
        \State Get mask $\mathscr{M}$ by elastic deforming and blurring $\mathcal{H}$.
        \State Get $X^g$ by blending $X^t$ with $X^s$ with $\mathscr{M}$.
    \end{algorithmic}
    \vspace{-2pt}
    \label{algo:data_augmentation}
\end{algorithm}

The workflow of \ourGeneratorAbbr~is summarized in Alg.~\ref{algo:data_augmentation} and illustrated in Fig.~\ref{fig:data_augmentation}.
Given a target video frame $X^t$, we take its $68$-point facial landmarks and retrieve another frame from different videos with different identities, so that the faces in the two frames have similar landmarks measured in $\ell_2$ norm.
For a pair of images, we first align their faces with the pre-computed landmarks and then detect the facial region by taking the convex hull of the landmarks.
Elastic deformation~\cite{ronneberger-icm15} is also employed to morph the convex hull:
we generate the smooth deformations using random displacement vectors sampled from a Gaussian distribution with a standard deviation of $6$ to $12$ pixels on a coarse $4$ by $4$ grid, and compute per-pixel displacements using bi-cubic interpolation.
The deformed mask is further blurred by a Gaussian kernel of size $16$.
Finally, the facial region of the source frame within the mask is stitched to the target frame using various blending methods~\cite{perez-acm03,naruniec-egsr20}. \ourGeneratorAbbr~outputs a forged video frame and the corresponding mask $\mathscr{M}$.

\section{Experiments}
\label{sec:exp}

We evaluated the performance of our approach (\ourModelAbbr~+ \ourGeneratorAbbr) against multiple state-of-the-art methods on seven publicly-available datasets.
First, we showed that our model achieves convincing performance under the in-dataset setting, where training and testing are conducted on the same dataset.
To demonstrate the superior generalization ability of our model, we conducted the cross-dataset evaluation by training the model with only \ourGeneratorAbbr-augmented real videos and testing on unseen datasets.
The ablation studies explored the contribution of each component in our model, such as the effect of \ourModelAbbr~and \ourGeneratorAbbr.

\begin{table*}
\centering
\small
\begin{tabular}{lcccccc|c}
\toprule
\multirow{2}{*}{Method} & \multirow{2}{*}{Backbone}
& \multirow{2}{*}{Train Set} & \multicolumn{5}{c}{Test Set (AUC (\%))} \\
\cmidrule(lr){4-8}
& & & DF & F2F & FS & NT & FF++ \\
\midrule
MIL~\cite{wang2018revisiting} & Xception & FF++ & 99.51 & 98.59 & 94.86 & 97.96 & 97.73 \\
Fakespotter~\cite{wang-arxiv19} & ResNet-50 & FF++, CD2, DFDC & - & - & - & - & 98.50 \\
XN-avg~\cite{roessler-iccv19} & Xception & FF++ & 99.38 & 99.53 & 99.36 & 97.29 & 98.89 \\
Face X-ray~\cite{li-cvpr20oral} & HRNet & FF++ & 99.12 & 99.31 & 99.09 & 99.27 & 99.20 \\
S-MIL-T~\cite{li-acm20} & Xception & FF++ & 99.84 & 99.34 & 99.61 & 98.85 & 99.41 \\
\midrule
\ourModelAbbr~+ \ourGeneratorAbbr~& ResNet-34 & FF++ & \textbf{100.00} & \textbf{99.57} & \textbf{100.00} & \textbf{99.58} & \textbf{99.79} \\
\bottomrule
\end{tabular}
\caption{\textit{In-dataset evaluation results on FF++.} Our method performs better on all manipulation types with a smaller backbone.}
\vspace{-5pt}
\label{tbl:in-dataset:ff++}
\end{table*}

\begin{table}[ht]
\centering
\begin{adjustbox}{width=0.475\textwidth}
\begin{tabular}{lcccc}
\toprule
\multirow{2}{*}{Method} & \multirow{2}{*}{Backbone} & \multirow{2}{*}{Train Set} & \multicolumn{1}{c}{Test Set (AUC (\%))} \\
\cmidrule(lr){4-4}
& & & CD2 \\
\midrule
Fakespotter~\cite{wang-arxiv19} & ResNet-50 & CD2 & 66.80 \\
Tolosana~\etal~\cite{tolosana-arxiv20} & Xception & CD2 & 83.60 \\
S-MIL-T~\cite{li-acm20} & Xception & CD2 & 98.84 \\
\midrule
\ourModelAbbr~+ \ourGeneratorAbbr~& ResNet-34 & CD2 & \textbf{99.98} \\
\bottomrule
\end{tabular}
\end{adjustbox}
\caption{\textit{In-dataset evaluation results on CD2.} We achieve saturated performance in terms of AUC.}
\vspace{-5pt}
\label{tbl:in-dataset:cd2}
\end{table}

\begin{table}[h]
\centering
\begin{adjustbox}{width=0.475\textwidth}
\begin{tabular}{lcccc}
\toprule
\multirow{2}{*}{Method} & \multirow{2}{*}{Backbone} & \multirow{2}{*}{Train Set} & \multicolumn{1}{c}{Test Set (AUC (\%))} \\
\cmidrule(lr){4-4}
& & & DFDC-P \\
\midrule
Tolosana~\etal~\cite{tolosana-arxiv20} & Xception & DFDC-P & 91.10 \\
S-MIL-T~\cite{li-acm20} & Xception & DFDC-P & 85.11 \\
\midrule
\ourModelAbbr~+ \ourGeneratorAbbr~& ResNet-34 & DFDC-P & \textbf{94.38} \\
\bottomrule
\end{tabular}
\end{adjustbox}
\caption{\textit{In-dataset evaluation results on DFDC-P.} Our method improves the best existing result by 3.28\% in terms of AUC.}
\vspace{-5pt}
\label{tbl:in-dataset:DFDC-P}
\end{table}

\subsection{Implementation Details}
\label{sec:exp:implementation_details}

\vspace{-5pt} \xhdr{Pre-processing.}
For each raw video frame, face crops are detected and tracked by using~\cite{li2020smot} and landmarks are detected by public toolbox~\cite{bulat-iccv17}.
We normalize all face crops with ImageNet mean $[0.485, 0.456, 0.406]$ and standard deviation $[0.229, 0.224, 0.225]$, and resize them to $256$\texttimes$256$.
We also use standard data augmentations, including
JPEG compression, Gaussian noise/blur, brightness contrast, random erasing, and color jittering.

\xhdr{Network Architecture.}
We adopt ResNet-34~\cite{he-cvpr16}
as backbone and initialized with pretrained weights on ImageNet~\cite{deng-cvpr09}.
Given a pre-processed video frame $X$ of size $H$\texttimes$W$\texttimes$3$, we first feed it into the backbone and extract the features $F$ after the \texttt{conv3} layer of size $H'$\texttimes$W'$\texttimes$256$, where $H'$=$H / 16$ and $W'$=$W / 16$.
Here each patch corresponds to a $16$\texttimes$16$ region in the original image.

\xhdr{Training.}
For each epoch, we randomly sample $32$ frames from every video, and the total number of training samples from $K$ videos is $32$\texttimes$K$.
The model is trained for $150$ epochs using Adam optimizer~\cite{kingma-iclr15} with batch size $128$, betas $0.9$ and $0.999$, and epsilon $10^{-8}$.
The learning rate is linearly increased from $0$ to $5$\texttimes$10^{-5}$ in the first quarter of the training iterations and is decayed to zero in the last quarter.
The hyper-parameter $\lambda$ is set to be $10$ by default.

\subsection{Settings}
\label{sec:exp:settings}

\vspace{-5pt} \xhdr{Training Data.}
Each of our training samples is of the form $(X,\textbf{V},y)$, where $X$ is the input video frame, $\textbf{V}$ is the ground truth 4D consistency volume, and $y$ is the binary label.
For a real frame,  $\textbf{V}$ is a 4D tensor of ones, indicating the image is self-consistent, and $y$ is zero.
There are two types of forged samples in our settings.
The first type is the deepfake video frame from the existing deepfake datasets, for which we find the corresponding real video frame and compute the structural dissimilarity (DSSIM)~\cite{wang-ip04} between them.
The mask $\mathscr{M}$ is then generated by taking a Gaussian blur of DSSIM followed by thresholding.
$\textbf{V}$ is computed from $\mathscr{M}$ by Eq.~\ref{eq:heatmap_groundtruth} and $y$ is one.
The second type is the fake image \textbf{augmented by \ourGeneratorAbbr} on real images, where $\textbf{V}$ is computed with the mask from \ourGeneratorAbbr~and $y$ is one.
By training with \ourGeneratorAbbr-augmented datasets, whenever a real data $(X^t,\bm{1},0)$ is sampled during the training, there is a $50\%$ chance that it is dynamically transformed by \ourGeneratorAbbr~into a fake data $(X^g,\textbf{V},1)$ where $X^g$ and $\mathscr{M}$ are the outputs of \ourGeneratorAbbr~as described in Alg.~\ref{algo:data_augmentation} and $\textbf{V}$ is computed with $\mathscr{M}$ by Eq.~\ref{eq:heatmap_groundtruth}.

More specifically, in the \textbf{in-dataset experiments}, our train set includes both the pristine and deepfake videos from the train split of the dataset (to be evaluated). Unless otherwise noted, we also utilize the fake samples augmented by \ourGeneratorAbbr~as data augmentation.
In the \textbf{cross-dataset experiments}, we follow prior work~\cite{li-cvpr20oral} and train with \textit{only real videos} from the raw version of FaceForensics++ (FF++)~\cite{roessler-iccv19}, augmented by \ourGeneratorAbbr.
Note that the cross-dataset setting is more close to the real-world scenarios where the potential attack types are not aware.

\xhdr{Test Data.}
FaceForensics++ (FF++)~\cite{roessler-iccv19} is by far the most popular benchmark for deepfake detection.
Its raw version contains $700$ videos for testing, including $140$ pristine and $560$ fake videos from $4$ different algorithms, which are Deepfakes (DF)~\cite{deepfakes}, Face2Face (F2F)~\cite{thies-cvpr16}, FaceSwap (FS)~\cite{kowalski-19} and NeuralTextures (NT)~\cite{thies-acm19}.
DeepfakeDetection (DFD)~\cite{dufour-dfd} dataset is released 
incorporated with FF++, supporting the deepfake detection research.
Celeb-DF-v1 (CD1) \& -v2 (CD2)~\cite{li-cvpr20_2} datasets consist of high-quality forged celebrity videos using advanced synthesis process.
Deepfake Detection Challenge (DFDC)~\cite{dolhansky-arxiv20} public test set is released for the Deepfake Detection Challenge, and DFDC Preview (DFDC-P)~\cite{dolhansky-arxiv19} is its preliminary version.
DFDC and DFDC-P contain many extremely low-quality videos, making them exceptionally challenging.
DeeperForensics-1.0 (DFR)~\cite{jiang-cvpr20} modifies the pristine videos in FF++ with new face IDs and more advanced techniques. More detailed statistics are provided in the Appendix.

\xhdr{Evaluation Metrics.}
We report the deepfake detection results with the most commonly used metrics in the literature, including the area under the ROC curve (AUC) and average precision (AP).
A higher AUC or AP value indicates better performance. 
To provide a comprehensive benchmark for future work, we report our performance on all datasets in terms of AUC, AP, as well as Equal Error Rate (EER) in the Appendix.
Unless otherwise noted, the evaluation results in the experiments are at video-level, computed by averaging the classification scores of the video frames.

\subsection{In-Dataset Evaluation}
\label{sec:exp:in-dataset}

In-dataset evaluation is abundantly adopted in the literature, where the focus is on specialization but not generalization.
To compare against the existing work, we consider three of the most popular datasets, which are FF++, CD2, and DFDC-P.
Given a dataset, our model is trained on both real and deepfake data from train split, and performance is evaluated with the corresponding test set.

The results for FF++, CD2, DFDC-P are shown in Table~\ref{tbl:in-dataset:ff++}, Table~\ref{tbl:in-dataset:cd2}, Table~\ref{tbl:in-dataset:DFDC-P}, respectively.
On average, compared to the state of the art, our approach improves the AUC score on these three datasets, from 96.45\% to 98.05\%.
Our models achieve the state of the art with near-perfect performance on CD2 ($99.98\%$) and all four manipulations of FF++ ($100.00\%$ on DF, $99.57\%$ on F2F, $100.00\%$ on FS, $99.58\%$ on NT), surpassing all existing work.
For DFDC-P, our model outperforms the state-of-the-art result by $3.28\%$ in terms of AUC score.
Note that the results reported for DFDC-P are comparably lower, due to the fact that a non-negligible portion of the dataset is of extremely low quality, \eg, human faces in some videos are hardly recognizable.
We also compare with prior work in terms of frame-level AUC on CD2, and outperform 
the state of the art~\cite{masi-eccv2020} by $8\%$ (see Appendix for more details).

\begin{table*}[t]
\small
\centering
\begin{tabular}{lcccccc|c|c}
\toprule
\multirow{2}{*}{Method} & \multirow{2}{*}{Backbone}
& \multirow{2}{*}{Train Set} & \multicolumn{6}{c}{Test Set (AUC (\%))} \\
\cmidrule(lr){4-9}
& & & DF & F2F & FS & NT & FF++ & DFD\\
\midrule
Face X-ray~\cite{li-cvpr20oral} & HRNet & FF++ (real data)
& 99.17 & 98.57 & 98.21 & \textbf{98.13} & 98.52 & 93.47 \\
\midrule
\ourModelAbbr~+ \ourGeneratorAbbr~& ResNet-34 & FF++ (real data)
& \textbf{100.00} & \textbf{98.97} & \textbf{99.86} & 97.63 & \textbf{99.11} & \textbf{99.07} \\
\bottomrule
\end{tabular}
\caption{\textit{Cross-dataset evaluation results on FF++ and DFD.} Our model is on par with Face X-ray~\cite{li-cvpr20oral} on FF++, but has better performance on DFD by 5.67\% in terms of AUC, with fewer network parameters.}
\vspace{-5pt}
\label{tbl:cross-dataset:ff++}
\end{table*}

\begin{table*}[h]
\small
\centering
\begin{tabular}{lcccccccc}
\toprule
\multirow{2}{*}{Method} & \multirow{2}{*}{Backbone}
& \multirow{2}{*}{Train Set} & \multicolumn{5}{c}{Test Set (AUC (\%))} \\
\cmidrule(lr){4-8}
& & & DFR & CD1 & CD2 & DFDC & DFDC-P\\
\midrule
Dang~\etal~\cite{dang-cvpr20} & Xception + Reg. & UADFV~\cite{yang-icassp18}, DFFD~\cite{dang-cvpr20} & - & 71.20 & - & - & - \\
DSP-FWA~\cite{li-cvprw19} & ResNet-50 & FF++ & - & - &  69.30 & - & - \\
Xception~\cite{roessler-iccv19} & Xception & FF++ & - & - & 73.04 & - & - \\
Masi~\etal~\cite{masi-eccv2020} & LSTM & FF++ & - & - & 76.65 & - & - \\
Face X-ray~\cite{li-cvpr20oral} & HRNet & FF++ & - & 80.58 & - & - & \textbf{80.92} \\
\midrule
\ourModelAbbr~+ \ourGeneratorAbbr~& ResNet-34 & FF++ (real data) & \textbf{99.41} & \textbf{98.30} & \textbf{90.03} & \textbf{67.52} & 74.37 \\
\bottomrule
\end{tabular}
\caption{\textit{Cross-dataset evaluation results on DFR, CD1, CD2, DFDC and DFDC-P datasets.} Our model out-performs the state of the art on CD1 and CD2, by about $18.00\%$ and $13.00\%$ in terms of AUC, and provides pioneering cross-dataset baselines on DFR (99.51\%) and DFDC (67.52\%). For DFDC-P, we have a lower AUC score but a higher AP score in comparison with the state of the art (see Table~\ref{tbl:cross-dataset:others:ap}).}
\vspace{-5pt}
\label{tbl:cross-dataset:others}
\end{table*}

\begin{table}[h]
\small
\centering
\begin{adjustbox}{max width=\linewidth}
\begin{tabular}{lcccccccc}
\toprule
\multirow{2}{*}{Method} & \multirow{2}{*}{Backbone}
& \multirow{2}{*}{Train Set} & \multicolumn{5}{c}{Test Set (AP (\%))} \\
\cmidrule(lr){4-8}
& & & CD1 & DFDC-P\\
\midrule
Face X-ray~\cite{li-cvpr20oral} & HRNet & FF++ & 73.33 & 72.65 \\
\midrule
\ourModelAbbr~+ \ourGeneratorAbbr~& ResNet-34 & FF++ (real data) & \textbf{98.97} & \textbf{82.94} \\
\bottomrule
\end{tabular}
\end{adjustbox}
\caption{\textit{Cross-dataset evaluation results on CD1 and DFDC-P datasets.} Our models can identify the attack video more precisely.
}
\vspace{-5pt}
\label{tbl:cross-dataset:others:ap}
\end{table}

\subsection{Cross-Dataset Evaluation}
\label{sec:exp:cross-dataset}

The generalization ability is an important indicator of the superiority of an algorithm. In the real world, the defense method cannot get any prior knowledge of the attacks. The cross-dataset evaluation is a widely-used approach to evaluate the generalization ability of an algorithm.

Table~\ref{tbl:cross-dataset:ff++} presents the cross-dataset evaluation results on FF++ and DFD, where we only used the real videos of FF++ for training.
The performance of our model is on par with Face X-ray~\cite{li-cvpr20oral} on FF++, achieving convincing results with more than $99.00\%$ in terms of AUC.
Interestingly, the performance gap between our model and Face X-ray~\cite{li-cvpr20oral} is much larger (99.07\% vs. 93.47\%) on DFD.
It is possible that the test data of FF++ is highly correlated with its training data, since they are very likely to be collected from the same source, whereas the correlation disappears in DFD.
The results demonstrate that predicting the source feature consistency can effectively generalize across different source cues, without overfitting to any spurious correlation among data from the same generation method.

We further evaluate our model on five more advanced datasets, as shown in Table~\ref{tbl:cross-dataset:others}.
In particular, our model outperforms the state of the art on CD1 and CD2, by about $18.00\%$ and $13.00\%$ in terms of AUC, and provides pioneering cross-dataset baselines on DFR (99.51\%) and DFDC (67.52\%).
On DFDC-P, our performance is comparable with Face X-ray~\cite{li-cvpr20oral}, where we get a lower AUC but a higher AP score, as shown in Table~\ref{tbl:cross-dataset:others:ap}. 
We compute the average AUC score among five out of seven datasets (except DFR and DFDC that has no published benchmarks) and find that our model's performance outperforms the state of the art (92.18\% vs. 86.03\%).
Meanwhile, we observe that both our model and state-of-art-art methods cannot achieve appealing results on DFDC/DFDC-P datasets, which motivates us to do failure analysis in Section~\ref{sec:exp:failure}~and~\ref{sec:exp:qualitative}.

\begin{figure*}[t]
\centering
\includegraphics[width=1.0\linewidth]{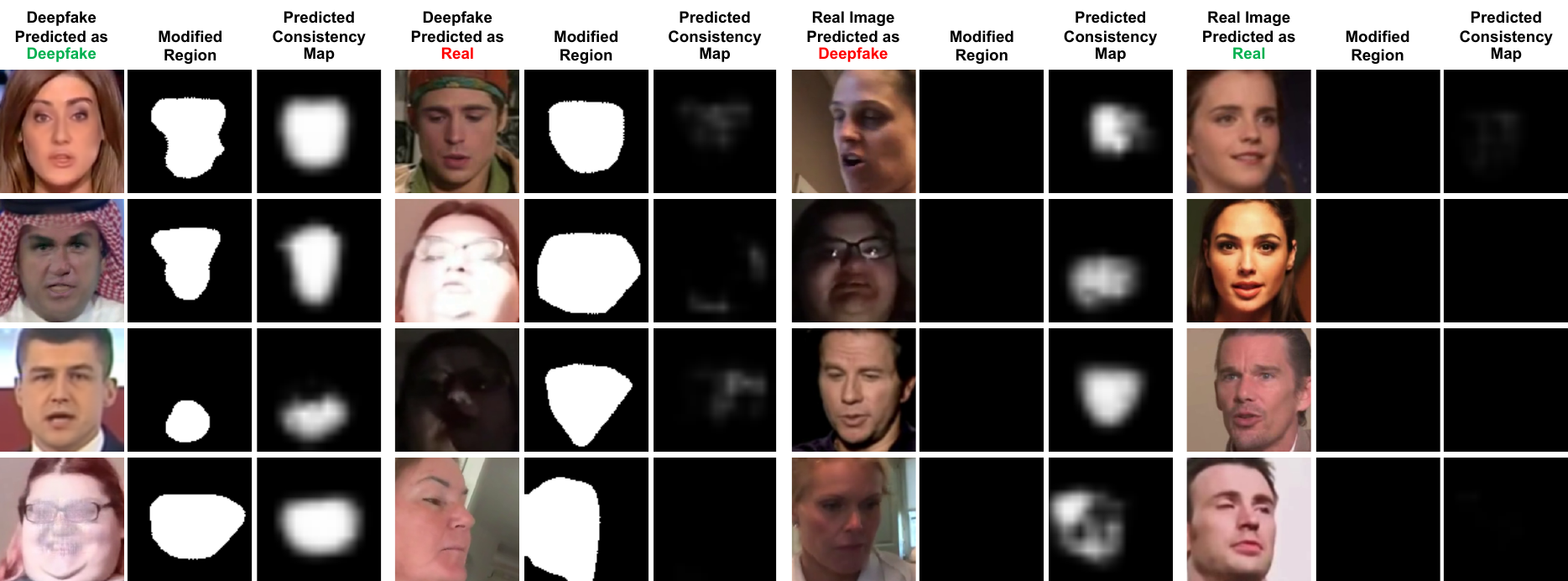}
\caption{\textit{Visualization of the predicted consistency maps $\widehat{\mathscr{M}}$, which try to localize the modified regions.} We use the model trained with real videos of FF++ augmented by \ourGeneratorAbbr~in the cross-dataset, and the predictions are computed from the predicted consistency volume, as mentioned in Section~\ref{sec:our_approach:pcl}. The ground truth modified regions are generated by DSSIM, as discussed in Section~\ref{sec:exp:settings}.}
\label{fig:qualitative_results}
\vspace{-5pt}
\end{figure*}

\subsection{Ablation Studies}
\label{sec:exp:ablation_studies}

\vspace{-5pt} \xhdr{Effect of \ourModelAbbr.}
We use $\lambda$ to balance between the consistency and classification losses, as shown in Eq.~\ref{eq:total_loss}.
By setting $\lambda=0$, we disable \ourModelAbbr~and get a network architecture equivalent to vanilla ResNet-34 with binary classification loss.
To exhibit the advantage of our consistency loss, we train models with increasing $\lambda$s and evaluate their cross-dataset generalizations with four test sets.
As shown in Table~\ref{tbl:ablation:consistency}, we follow the cross-dataset setting and train all models on real data of FF++ augmented by \ourGeneratorAbbr~and report the AUC score for performance comparison.
We observe that training with $\lambda>0$ significantly outperforms the training with $\lambda=0$.
Especially, the performance on DFDC is improved by $15.8\%$ in terms of AUC.
The results validate that it is beneficial to use large $\lambda$ during training, which also suggests that \ourModelAbbr~plays a dominant role in the success.

\xhdr{Effect of \ourGeneratorAbbr~as Joint-Training.}
We have been training with the consistency loss on datasets augmented by \ourGeneratorAbbr.
\ourGeneratorAbbr~generates fake data dynamically, enhancing the training data variety, thereby improving the performance and generalization.
To demonstrate the effect of \ourGeneratorAbbr~, we conduct ablation studies by training on either DF or DFDC-P and benchmarking on DFR, CD2, DFDC, and DFDC-P test sets.
Table~\ref{tbl:ablation:i2g} shows that models trained on data augmented by \ourGeneratorAbbr~outperform the straightforward combination of routine data augmentations and blending methods used in the baseline.
In particular, we train models on the DF with or without \ourGeneratorAbbr, whereas the performance of the latter is improved by an average AUC score of $7.18\%$ over four test sets.
The model trained with the DFDC-P train set has noticeably better performance on DFDC and DFDC-P test sets comparing to previous models, but generalizes poorly on other datasets such as DFR.
\ourGeneratorAbbr~improves model's generalization, for example, performance on DFR is raised from $51.61\%$ to $92.25\%$ in terms of AUC, with a minor sacrifice to the performance on DFDC-P.

\xhdr{Effect of \ourGeneratorAbbr~as Pre-Training.}
When even computing the DSSIM masks of deepfakes is not feasible, one can always use our consistency loss to \textit{pretrain} the model with any real data augmented by \ourGeneratorAbbr.
After that, any standard training for deepfake detection or related tasks can be conducted with the whole dataset.
In particular, we conducted an experiment where we first pretrained a ResNet-34 on the real data of DF augmented by \ourGeneratorAbbr~for consistency prediction, and finetuned with both real and fake data of DF for classification.
We report the evaluation results on DFR, CD2, DFDC, DFDC-P with $99.57\%$, $91.88\%$, $68.95\%$, $79.17\%$ ($84.89\%$ on average) in terms of AUC, respectively.
These results are unsurprisingly lower than those of joint-training, but still significantly outperform the baseline (79.19\% on average).
Besides, our pretrained models may not necessarily be trained from the established deepfake datasets.
\ourGeneratorAbbr~can be potentially applied to any face image or video datasets, such as IMDb-Face~\cite{wang2018devil} and YouTube Faces~\cite{wolf2011face}, providing a stronger pretrained model for deepfake-related research.

\xhdr{Choice of Patch Size.}
We evaluate the effectiveness of using different patch sizes.
Conceptually, larger patches get coarser consistency maps that may reduce their efficacy on forgery detection,
while smaller patches may not contain enough information of source features and induce extra computation cost.
In particular, we evaluate our cross-dataset model from Table~\ref{tbl:cross-dataset:ff++} with the patch size of $4$\texttimes$4$, $8$\texttimes$8$, $16$\texttimes$16$, and $32$\texttimes$32$ on FF++,
and get $98.32\%$, $98.35\%$, $99.11\%$, and $98.74\%$ in terms of AUC, respectively. 

\begin{table}[t]
\centering
\begin{adjustbox}{max width=\linewidth}
\begin{tabular}{lcccccc}
\toprule
\multirow{2}{*}{Method} & \multirow{2}{*}{\makecell{Hyper-\\Parameter}} & \multicolumn{4}{c}{Test Set (AUC (\%))} & \multirow{2}{*}{Avg} \\
\cmidrule(lr){3-6}
& & DFR & CD2 & DFDC & DFDC-P \\
\midrule
\ourGeneratorAbbr~& $\lambda=0$ & 95.12 & 78.18 & 51.72 & 69.93 & 73.74 \\
\midrule
\ourModelAbbr~+ \ourGeneratorAbbr~& $\lambda=1$ & 99.1 & 86.52 & 60.65 & 74.13 & 80.10 \\
\ourModelAbbr~+ \ourGeneratorAbbr~& $\lambda=10$ & 99.41 & 90.03 & \textbf{67.52} & \textbf{74.37} & \textbf{82.83} \\
\ourModelAbbr~+ \ourGeneratorAbbr~& $\lambda=100$ & \textbf{99.78} & \textbf{90.98} & 63.22 & 74.36 & 82.09 \\
\toprule
\end{tabular}
\end{adjustbox}
\caption{\textit{Ablation study on the effect of \ourModelAbbr~on DFR, CD2, DFDC, and DFDC-P datasets.} The use of large $\lambda$ significantly improves the cross-dataset performance, especially on DFDC.}
\vspace{-5pt}
\label{tbl:ablation:consistency}
\end{table}

\begin{table}[t]
\centering
\begin{adjustbox}{max width=0.475\textwidth}
\begin{tabular}{lccccccc}
\toprule
\multirow{2}{*}{Method} & \multirow{2}{*}{Train Set}
& \multicolumn{4}{c}{Test Set (AUC (\%))} & \multirow{2}{*}{Avg} \\
\cmidrule(lr){3-6}
& & DFR & CD2 & DFDC & DFDC-P \\
\midrule
\ourModelAbbr & DF & 90.42 & 84.59 & 66.26 & 75.49 & 79.19 \\
\ourModelAbbr~+ \ourGeneratorAbbr~& DF & \textbf{99.64} & \textbf{91.92} & \textbf{73.08} & \textbf{80.83} & \textbf{86.37} \\
\midrule
\ourModelAbbr & DFDC-P & 51.61 & 82.82 & 69.14 & \textbf{95.53} & 74.78 \\
\ourModelAbbr~+ \ourGeneratorAbbr~& DFDC-P & \textbf{92.25} & \textbf{87.65} & \textbf{71.12} & 94.38 & \textbf{86.35} \\
\bottomrule
\end{tabular}
\end{adjustbox}
\caption{\textit{Ablation study on the effect of \ourGeneratorAbbr~as joint-training on DF and DFDC-P datasets.} \ourGeneratorAbbr~can enhance the variety of training data, thereby improving the generalization of our model.}
\vspace{-5pt}
\label{tbl:ablation:i2g}
\end{table}

\subsection{Qualitative Results}
\label{sec:exp:qualitative}
\ourModelAbbr~not only improves the representation learning for deepfake detection, but also can be used to generate the interpretable visualizations clues (Sec.~\ref{sec:our_approach:pcl}) about the modified region.
Figure~\ref{fig:qualitative_results} visualizes some examples generated by~\ourModelAbbr~along with the corresponding input images and ground truth.
When feeding a real image, in most cases, the visualization is a pure blank image, indicating that the input's source features are consistent.
When testing the deepfakes, the predicted consistency map can adequately match with the ground truth. 
We also compute the average value of the consistency volume from real images, and get $0.9854$ and $0.9866$ using in- and cross-dataset models, respectively.
These statistical numbers indicate that \ourModelAbbr~predicts all entries in the consistency volume of correctly predicted samples to be consistent with high average confidence, rather than simply segmenting the full face region.
We also investigate some failure cases, the inconsistencies are caught by mistake in our model, which might be caused by the lighting and unusual texture.
Besides, we observe that lower quality samples lead to false-negative predictions due to either high compression or high/low exposure.

\subsection{Limitations}
\label{sec:exp:failure}
Although our results are encouraging, our approaches still have limitations, which raise opportunities for future work.
First, as the game between the forger and the detector is an arms race, one can expect the cues that any published detection method relies on to be removed with the best efforts in the near future.
For example, entire face synthesis trains a generative model that directly outputs the whole image, which should be self-consistent by our hypothesis; it is unknown if \ourModelAbbr~can handle this type of face forgery.
Second, as the false prediction samples indicated, our model can be further improved on low-quality data.

\section{Conclusion}

We proposed \ourModelName~(\ourModelAbbr) to detect face forgeries generated by stitching-based techniques and localize the manipulated regions, based on a less attended cue: the inconsistency of source features within the modified images.
\ourModelAbbr~only contains a few parameters and can serve as a plugin module upon common backbone networks.
We also developed a new light-weight image synthesis method, called \ourGeneratorName~(\ourGeneratorAbbr), to efficiently support \ourModelAbbr~training by dynamically generating forged images along with annotations of their manipulated regions.
Experimental results showed that \ourModelAbbr~and \ourGeneratorAbbr~are competitive against state-of-the-art methods on seven popular datasets, providing a strong baseline for future research.

\section{Appendix}

\subsection{Video-Level Comprehensive Results}
Due to the space limit, we mainly reported the performance of our models in term of AUC in the main text.
Here we provide more results in terms of AUC, AP, and EER for Table~\ref{tbl:cross-dataset:ff++} and~\ref{tbl:cross-dataset:others} in the main text.
Note that these results are at \textit{video-level}, computed by averaging the classification scores of all video frames.
The experiments are conducted under the \textit{cross-dataset} setting, in which we train our model only with real data from the raw version of FF++~\cite{roessler-iccv19} and the fake/positive samples are generated by \ourGeneratorAbbr~(more detailed are included in Sec.~\ref{sec:exp:settings}).

\vspace{-1mm}

\begin{table}[H]
\centering
\begin{adjustbox}{max width=0.90\linewidth}
\begin{tabular}{lcccccc|c|c}
\toprule
\multirow{2}{*}{Method} & \multirow{2}{*}{Test Set} & \multicolumn{3}{c}{Evaluation Metrics (\%)} \\
\cmidrule(lr){3-5}
& & AUC & AP & EER \\
\midrule
\multirow{11}{*}{\ourModelAbbr~+ \ourGeneratorAbbr}
& DF~\cite{deepfakes}             & 100.00 & 100.00 &  0.00 \\
& F2F~\cite{thies-cvpr16}         & 98.97  & 99.32  &  3.57 \\
& FS~\cite{kowalski-19}           & 99.86  & 99.86  &  1.43 \\
& NT~\cite{thies-acm19}           & 97.63  & 98.20  &  6.43 \\
& FF++~\cite{roessler-iccv19}     & 99.11  & 99.80  &  3.57 \\
\cmidrule(lr){2-5}
& DFD~\cite{dufour-dfd}           & 99.07  & 99.89  &  4.42 \\
& DFR~\cite{jiang-cvpr20}         & 99.41  & 99.51  &  3.48 \\
& CD1~\cite{li-cvpr20_2}          & 98.30  & 98.97  &  7.89 \\
& CD2~\cite{li-cvpr20_2}          & 90.03  & 94.45  & 17.98 \\
& DFDC~\cite{dolhansky-arxiv20}   & 67.52  & 69.99  & 37.18 \\
& DFDC-P~\cite{dolhansky-arxiv19} & 74.37  & 82.94  & 31.87 \\
\bottomrule
\end{tabular}
\end{adjustbox}
\caption{\textit{Comprehensive evaluation results of our model in terms of video-level AUC, AP, and EER on seven datasets.}}
\label{tbl:cross-dataset:full}
\end{table}

\vspace{-3mm}

\begin{figure*}
\centering
\includegraphics[width=1.0\linewidth]{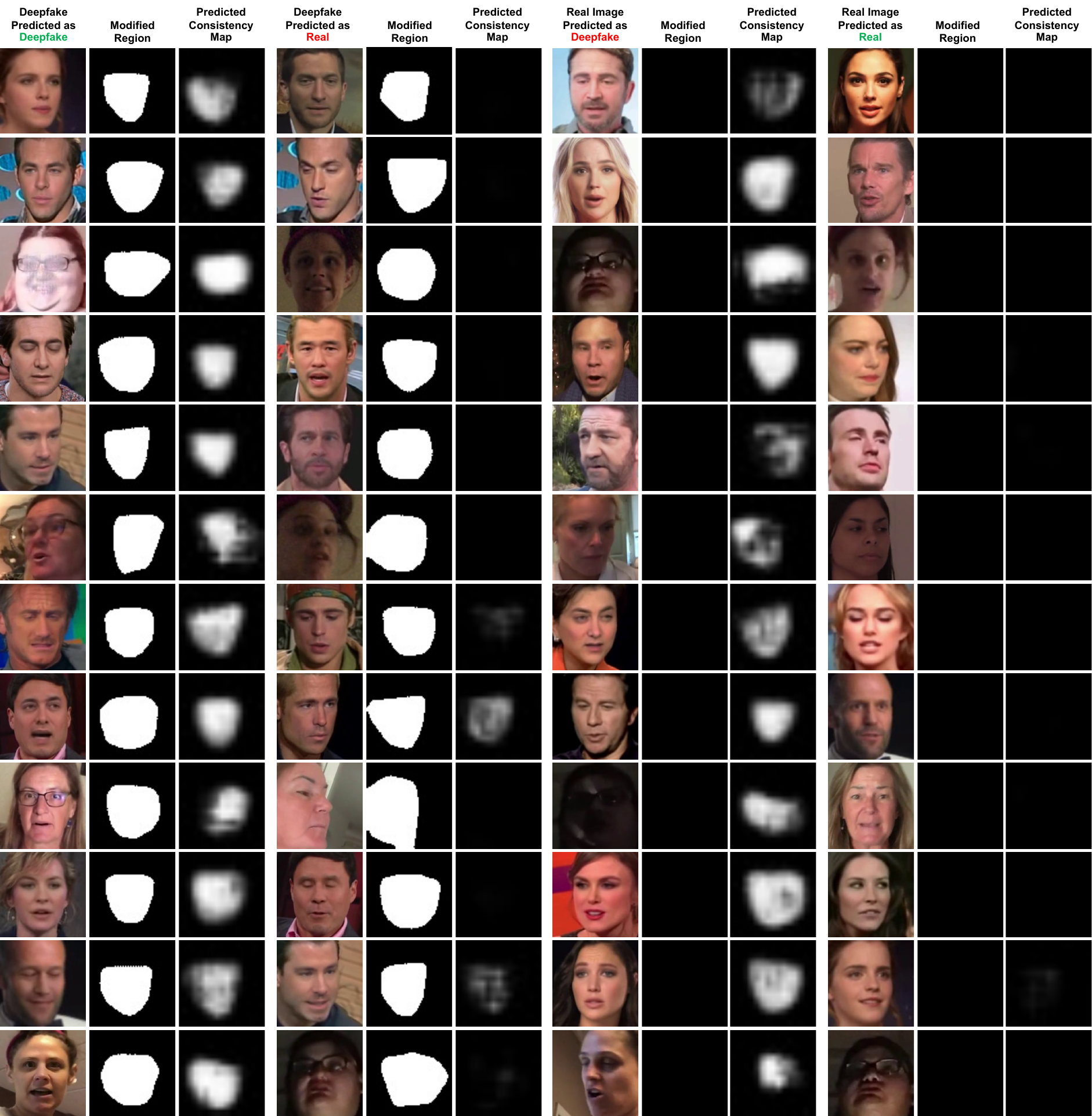}
\caption{
\textit{Visualization of the predicted consistency maps $\widehat{\mathscr{M}}$, which try to localize the modified regions.} We use the model trained with real videos of FF++ augmented by I2G in the cross-dataset, and the predictions are computed from the predicted consistency volume, as mentioned in Section~{\color{red}3.1}. The ground truth modified regions are generated by DSSIM, as discussed in Section~{\color{red}4.2}.
}
\vspace{100pt}
\label{fig:qualitative_results_2}
\end{figure*}

\xhdr{Additional qualitative results} are shown in Fig.~\ref{fig:qualitative_results_2}. These images are randomly chosen from CD2~\cite{li-cvpr20_2} and DFDC~\cite{dolhansky-arxiv20} test sets, which are currently the most challenging datasets in deepfake detection.
The visualization is obtained by up-sampling the 2D global heatmap $\widehat{\mathcal{M}}$ to the size of $H$\texttimes$W$ to match the input size, where $\widehat{\mathcal{M}}$ is generated by fusing the predicted 4D consistency volume $\widehat{\textbf{V}}$.

\subsection{Frame-Level Results on Celeb-DF-v2}
Deepfake detection accuracy is usually reported at the video-level. Methods aggregate the frame-level scores to form video-level predictions using various strategies, \eg, averaging (ours), confident strategy~\cite{seferbekov-20}, LSTM~\cite{guera-avss18}.
Here we compare our model with other state-of-the-art methods in terms of frame-level AUC on CD2~\cite{li-cvpr20_2}.
All models are trained under the \textit{cross-dataset} setting (see Sec.~{\color{red}4.2} for more details). This means they are trained on FF++ and evaluated on CD2. As shown in~Table~\ref{tbl:cross-domain:frame-level:cd2}, our model outperforms other state-of-the-art method~\cite{masi-eccv2020} by over $8\%$. Note the baseline results are directly cited from Masi~\etal~\cite{masi-eccv2020}. Different compression levels of FF++ are adopted by the methods, \eg, c23 for Zhao~\etal~\cite{zhao-cvpr21}, c40 for Masi~\etal~\cite{masi-eccv2020}, and raw for ours.

\begin{table}[H]
\centering
\begin{adjustbox}{max width=0.90\linewidth}
\begin{tabular}{l|c}
\toprule
Method & CD2 (Frame-Level AUC (\%)) \\
\midrule
Two-stream~\cite{zhou-cvprw17} & 53.8 \\
Meso4~\cite{afchar-wifs18} & 54.8 \\
MesoInception4 & 53.6 \\
HeadPose~\cite{yang-icassp18} & 54.6 \\
FWA~\cite{li-cvprw19} & 56.9 \\
VA-MLP~\cite{matern2019exploiting} & 55.0 \\
VA-LogReg & 55.1 \\
Xception-raw~\cite{roessler-iccv19} & 48.2 \\
Xception-c23 & 65.3 \\
Xception-c40 & 65.5 \\
Multi-task~\cite{nguyen-btas19} & 54.3 \\
Capsule~\cite{nguyen2019capsule} & 57.5 \\
DSP-FWA~\cite{li-cvprw19} & 64.6 \\
Zhao~\etal~\cite{zhao-cvpr21} & 67.4 \\
Masi~\etal~\cite{masi-eccv2020} & 73.4 \\
\midrule
\ourModelAbbr~+ \ourGeneratorAbbr~ & \textbf{81.8} \\
\toprule
\end{tabular}
\end{adjustbox}
\caption{
\textit{
Cross-dataset evaluation results of our model
in terms of frame-level AUC on CD2 dataset.}
The performances of existing methods are cited for comparison.
}
\label{tbl:cross-domain:frame-level:cd2}
\end{table}

\subsection{Computational Complexity}
Each vector of size $C$ in the source feature map of size $H/P$\texttimes$W/P$\texttimes$C$  corresponds to a $P$\texttimes$P$ patch in the input image, and $P$ is the down-sampling factor.
The additional computations from \ourModelAbbr~consist of (1) embedding feature map into size $H/P$\texttimes$W/P$\texttimes$C'$ with $O(CC'HW/P^2)$ flops and (2) computing pair-wise consistency with $O(C'H^2W^2/P^4)$ flops.
In practice,
we use $H$=$W$=256, $C$=$256$, $C'$=$128$, $P$=$16$, and ResNet-34 as backbone, and \ourModelAbbr~contributes $48.12$M FLOPs, $65.5$K parameters and $0.0009$ seconds to a total of $9.62$G FLOPs, $21.3$M parameters and $0.0234$ seconds, for a forward pass of a single image,
running on one NVIDIA Tesla V100 GPU with 16GB of memory.

\vspace{5pt}
\subsection{Consistency Map Generation}
For visualization, we assume there are two source feature groups in each image. The top-left image patch belongs to group 0 and dissimilar patches to it belong to group 1. For a patch at $(h, w)$ with a $H$\texttimes$W$ consistency score matrix $\widehat{M}^{\mathcal{P}_{h,w}} = \{ \widehat{m}^{\mathcal{P}_{h,w}}_{i,j} \}$, we get a soft group assignment as its cosine similarity to the top-left patch, $\widehat{m}^{\mathcal{P}_{h,w}}_{0,0}$. We calculate the corresponding grayscale value of this patch in the consistency map by averaging $|\widehat{m}^{\mathcal{P}_{h,w}}_{0,0} - \widehat{m}^{\mathcal{P}_{h,w}}_{i,j}|$.
In this way, similar grayscale values in the visualization indicate the corresponding patches are likely to belong to the same source feature group. 

\newpage
\subsection{Different Backbones}
\begin{figure}[H]
\centering
\includegraphics[width=0.95\linewidth]{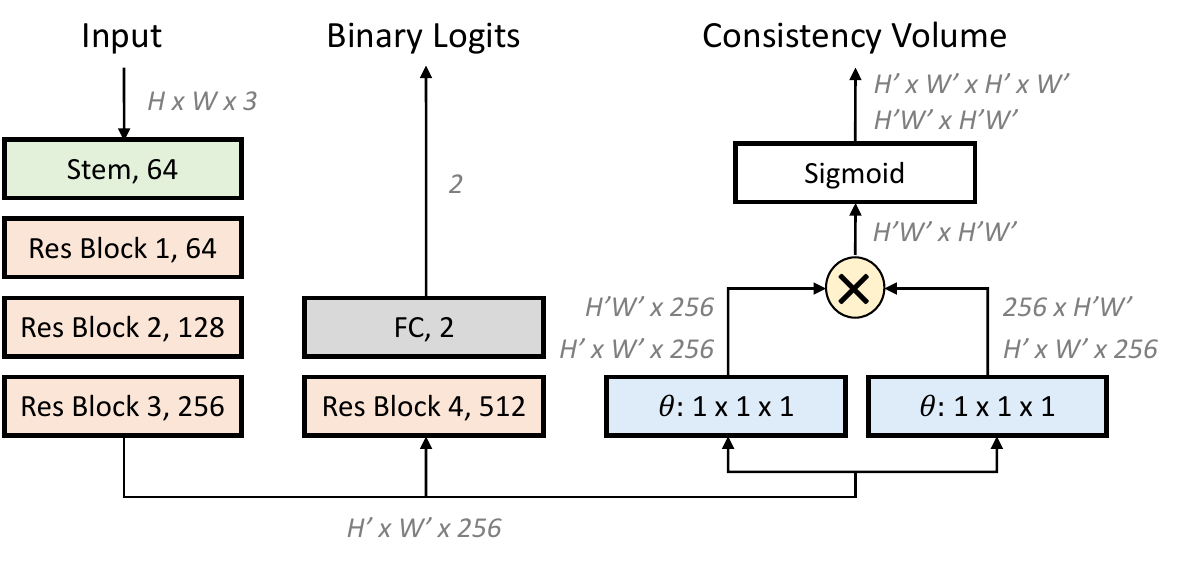}
\caption{
\textit{An example of \ourModelAbbr~architecture,} where ResNet-34
is adopted as backbone.
The features are shown as the shape of their tensors,
and proper reshaping is performed. $\bigotimes$ denotes matrix multiplication, and $\theta$ is
$1$\texttimes$1$\texttimes$1$
convolution.
}
\label{fig:detailed_network}
\vspace{-10pt}
\end{figure}

In this paper, we adopt ResNets as backbone, as they are among the most popular classification networks; an example of the \ourModelAbbr~architecture is illustrated in Fig.~\ref{fig:detailed_network}.
Our contribution does not rely on any particular choice of the model architecture.
However, it is still interesting to discover if increasing the model capacity improves the cross-dataset generalization.
We build our model with ResNets of various depths, and report the results in Table~\ref{tbl:ablation_study_depth}.
The performance increase is noticeable from ResNet-18 to ResNet-50 but diminishes as we go deeper.

\vspace{-3mm}

\begin{table}[H]
\centering
\begin{adjustbox}{max width=\linewidth}
\begin{tabular}{lccccc}
\toprule
\multirow{2}{*}{Backbone} & \multicolumn{4}{c}{Test Set (AUC (\%))} & \multirow{2}{*}{Avg} \\
\cmidrule(lr){2-5}
& DFR & CD2 & DFDC & DFDC-P \\
\midrule
ResNet-18
& 96.92 & 79.59 & 58.22 & 68.23 & 75.74\\
ResNet-34
& 99.41 & 90.03 & 67.52 & 74.37 & 82.83 \\
ResNet-50
& 99.13 & 90.70 & \textbf{70.69} & \textbf{75.10} & \textbf{83.90} \\
ResNet-152
& \textbf{99.5} & \textbf{90.88} & 70.42 & 69.77 & 82.64\\
\toprule
\end{tabular}
\end{adjustbox}
\caption{\textit{Ablation study of different backbones.} The models are all trained with $\lambda=10$. The performance saturates as the depth of the model increases.}
\label{tbl:ablation_study_depth}
\end{table}

\subsection{Test Set Statistics}
\vspace{-7pt}

\begin{table}[H]
\centering
\begin{tabular}{l|c}
\toprule
Test Set & \# Real / Fake Videos \\
\midrule
FaceForensics++ (FF++)~\cite{roessler-iccv19} & 140 / 560 \\
FF++ - Deepfakes (DF)~\cite{deepfakes} & 140 / 140 \\
FF++ - Face2Face (F2F)~\cite{thies-cvpr16} & 140 / 140 \\
FF++ - FaceSwap (FS)~\cite{kowalski-19} & 140 / 140 \\
FF++ - NeuralTextures (NT)~\cite{thies-acm19} & 140 / 140 \\
DeepfakeDetection (DFD)~\cite{dufour-dfd} & 363 / 3431 \\
Celeb-DF-v1 (CD1)~\cite{li-cvpr20_2} & 38 / 62 \\
Celeb-DF-v2 (CD2)~\cite{li-cvpr20_2} & 178 / 340 \\
DFDC Public (DFDC)~\cite{dolhansky-arxiv20} & 2000 / 2000 \\
DFDC Preview (DFDC-P)~\cite{dolhansky-arxiv19} & 276 / 504 \\
DeeperForensics-1.0 (DFR)~\cite{jiang-cvpr20} & 201 / 201 \\
\toprule
\end{tabular}
\caption{\textit{Statistics of real and fake videos in the test sets.}}
\label{tbl:test_stats}
\end{table}

{\small
\bibliographystyle{ieee_fullname}
\bibliography{egbib}
}

\end{document}